\documentclass{article}

\usepackage{PRIMEarxiv}

\usepackage[utf8]{inputenc} 
\usepackage[T1]{fontenc}    
\usepackage{xcolor}
\usepackage{hyperref}       
\usepackage{url}            
\usepackage{booktabs}       
\usepackage{amsfonts}       
\usepackage{nicefrac}       
\usepackage{microtype}      
\usepackage{lipsum}
\usepackage{fancyhdr}       
\usepackage{graphicx}       
\graphicspath{{media/}}     

\usepackage{amsmath}
\usepackage{amsthm}
\usepackage{algorithm}
\usepackage{algorithmic}
\usepackage[toc,page]{appendix}
\usepackage[inline]{enumitem}

\title{\name: Advice Explanations in \\ Complex Repeated Decision-Making Environments
\thanks{\textit{\underline{Citation}}: This preprint has been accepted for publication at IJCAI'24. The final verson is available at \href{https://doi.org/10.24963/ijcai.2024/875}{10.24963/ijcai.2024/875}} 
}

\author{
  Sören Schleibaum \\
  Clausthal University of Technology \\
  \texttt{soeren.schleibaum@tu-clausthal.de} \\
  \And
  Lu Feng \\
  University of Virginia\\
  \texttt{lu.feng@virginia.edu} \\
  \And
  Sarit Kraus \\
  Bar-Ilan University\\
  \texttt{sarit@cs.biu.ac.il} \\
  \And
  Jörg P. Müller \\
  Clausthal University of Technology\\
  \texttt{joerg.mueller@tu-clausthal.de} \\
}

\newcommand\name{{\textsc{Adesse}}}

\DeclareMathOperator*{\argmax}{arg\,max}

\usepackage[english]{babel}
\usepackage{hyperref}
\addto\extrasenglish{
    
}
\addto\extrasenglish{
    
}

\begin{document}
\maketitle

\begin{abstract}
In the evolving landscape of human-centered AI, fostering a synergistic relationship between humans and AI agents in decision-making processes stands as a paramount challenge. This work considers a problem setup where an intelligent agent comprising a neural network-based prediction component and a deep reinforcement learning component provides advice to a human decision-maker in complex repeated decision-making environments. Whether the human decision-maker would follow the agent's advice depends on their beliefs and trust in the agent and on their understanding of the advice itself. To this end, we developed an approach named {\name} to generate explanations about the adviser agent to improve human trust and decision-making. Computational experiments on a range of environments with varying model sizes demonstrate the applicability and scalability of {\name}. Furthermore, an interactive game-based user study shows that participants were significantly more satisfied, achieved a higher reward in the game, and took less time to select an action when presented with explanations generated by {\name}. These findings illuminate the critical role of tailored, human-centered explanations in AI-assisted decision-making.
\end{abstract}


\section{Introduction} \label{sec:intro} 
\begin{figure}[t]
\centering
\includegraphics[width=.5\textwidth]{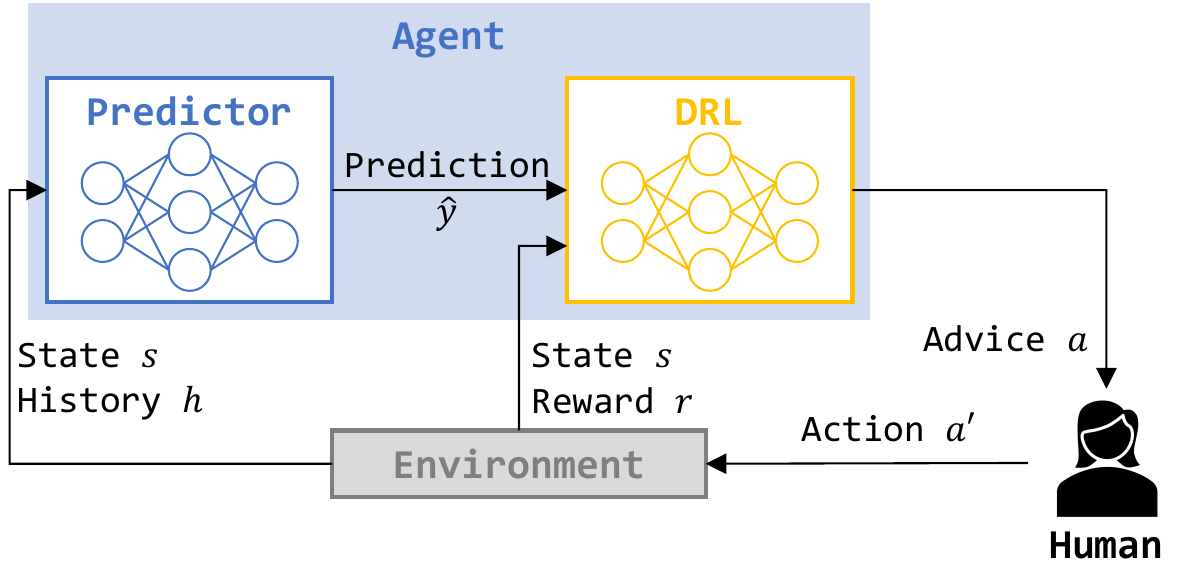}
\caption{An agent consisting of two components provides advice to a human decision-maker.}
\label{fig:problem}
\end{figure}

Making complex decisions repeatedly in a dynamic environment is very challenging for humans. An intelligent agent can support human decision-making by providing advice. 
We consider an adviser agent consisting of two components as shown in \autoref{fig:problem}. 
At each step, the agent first makes some predictions about the future, and then computes advice based on the prediction and the current state using deep reinforcement learning (DRL). 
Such adviser agents can and are being used in many real-world applications: For example, providing advice to police officers scheduled through place-based predictive policing~\cite{meijer2019predictive}, providing advice to taxi drivers based on the prediction of future pick-up requests from passengers~\cite{farazi2021deep}, or providing advice to firefighters based on the prediction of wildfire risk~\cite{julian2019distributed}.

Studies have found that the degree to which humans follow an intelligent agent's advice depends on their beliefs about the agent's performance on a given task~\cite{vodrahalli2022humans}, and that providing explanations improves humans' acceptance and trust in the agent's advice~\cite{zhang2020effect,shin2021exptrust}.
Hence, this work aims at generating explanations about the adviser agent to improve human's trust and decision-making. 

Existing methods for explaining AI-based systems mostly treat the entire system as a black-box model;
the generated explanations could be in different output formats (e.g., numerical, textual, visual), but each method usually only focuses on one type of explanations~\cite{adadi2018peeking,guidotti2018survey,speith2022review}. 
For example, there are several methods (e.g.,~\cite{ribeiro2016should,lundberg2017unified}) explaining the feature importance of prediction; and there is a growing body of research on explainable reinforcement learning~\cite{vouros2022explainable,wells2021explainable,heuillet2021explainability,puiutta2020explainable}. Nevertheless, to the best of our knowledge, none of the prior works generates explanations for both prediction and DRL.

In this work, we present a novel approach named {\name} (\underline{AD}vice \underline{E}xplanation\underline{S} in complex repeated deci\underline{S}ion-making \underline{E}nvironments)\footnote{{\name} means ``to aid'' in Latin.}. 
{\name} peeks inside the black-box model of an adviser agent, leveraging the agent's two-component structure to generate explanations with both textual and visual information. 
Specifically, an explanation generated by {\name} includes three key elements: 
(1) a short list of top-ranked input features that contribute the most to the agent's prediction; 
(2) a heatmap visualizing domain-specific indices summarizing the DRL input features;
and (3) arrows in various shades of gray overlaying the heatmap to illustrate a trained DRL policy with state importance. 

A key innovation of {\name} is to generate informative explanations that capture multiple aspects of the adviser agent, from the prediction input to the DRL input to the trained DRL policy.
Furthermore, {\name} reduces the explanation size via selecting top-ranked input features of the prediction and using domain-specific indices to succinctly explain DRL input features.  

We adopt LIME~\cite{ribeiro2016should}, a popular method for explaining black-box models, as a baseline for comparison.
LIME generates explanations represented as (multiple) saliency maps visualizing each input feature's influence on the agent's advice, which can be overwhelming when there is a large number of input features.  
We hypothesize that explanations generated by {\name} can be more effective in assisting human decision-making than the baseline. 

Computational experiments demonstrate that {\name} can be successfully applied to a range of environments and scales over varying model sizes. In all cases, {\name} generates smaller explanations using less time, compared with LIME. 

Additionally, we conduct an interactive game-based user study to evaluate the effectiveness of generated explanations. 
Study results show that participants were significantly more satisfied, achieved a higher reward in the game, and took less time to select an action when presented with explanations generated by {\name} rather than the baseline.

\section{Related Work} \label{sec:related} 

\subsection{Position within the XAI Literature}

The research field of \emph{explainable artificial intelligence} (XAI) has been growing rapidly in recent years, attracting increasing attention ~\cite{adadi2018peeking,guidotti2018survey,speith2022review,saeed2023explainable,anjomshoae2019explainable}. 
Here, we position this work based on a taxonomy of XAI methods described in~\cite{speith2022review}. 

First, depending on the stage when explanations are generated, there are \emph{ante-hoc} and \emph{post-hoc} methods.
This work belongs to the latter since {\name} generates explanations after the agent has been trained. 

Second, there are \emph{model-specific} and \emph{model-agnostic} methods. 
{\name} is agnostic to the underlying machine learning techniques for prediction and advice computation. 

Third, the scope of explanations can be \emph{global} or \emph{local}. 
An explanation generated by {\name} consists of three key elements, in which the first element (i.e., a list of top-ranked features for the prediction at a grid cell) is local and the other two elements (i.e., domain-specific indices and arrows for visualizing a DRL policy) are global. 

Moreover, XAI methods generate explanations in diverse output formats, including numerical, textual, visual, rules, models, etc.
{\name} generates explanations displayed visually as a heatmap together with textual information about a short list of top-ranked features.

Last but not least, the lack of user studies is a major limitation across many existing XAI works, as pointed out in several survey papers~\cite{wells2021explainable,kraus2020ai,chakraborti2020emerging}.
This work overcomes this limitation by adopting an interactive game-based user study for evaluation. 

At first glance, the motivating examples described in the next section seem similar to the task of goal recognition. However, in contrast to goal recognition (see \cite{shvo2020active}), we have time-dependent targets and do not learn a probability distribution over goals. Consequently, we cannot base our work on those explaining goal recognition, e.g. \cite{alshehri2023explainable}.

\subsection{Feature Importance}

Many XAI methods explain black-box models via computing \emph{feature importance} (e.g., how much a feature contributes to a prediction). 
\emph{Local Interpretable Model-agnostic Explanations} (LIME)~\cite{ribeiro2016should} and \emph{SHapley Additive exPlanations} (SHAP)~\cite{lundberg2017unified} are two of the most popular methods in this category. 

LIME focuses on training local surrogate models to explain individual predictions. 
This method works by first generating a new dataset comprising perturbed samples and the corresponding predictions of the black box model, and then using this new dataset to train an interpretable surrogate model that is weighted by the proximity of the sampled instances to the instance of interest. 
The learned surrogate model can provide a good approximation of local predictions, but does not necessarily guarantee global accuracy. 

On the other hand, SHAP computes Shapley values of features (i.e., the average marginal contribution of a feature value across all possible coalitions) by considering all possible predictions for an instance using all possible combinations of inputs.
Because of this exhaustive analysis, SHAP can take much longer computation time than LIME. 
The authors of \cite{lundberg2017unified} show that SHAP can guarantee properties such as accuracy and consistency, while LIME is a subset of SHAP but lacks these properties.

\subsection{Explainable Reinforcement Learning}

\emph{Explainable reinforcement learning} (XRL) has emerged as a sub-field of XAI with a growing body of research~\cite{vouros2022explainable,wells2021explainable,heuillet2021explainability,puiutta2020explainable}.
Existing XRL methods can be distinguished by the scope of explanations. Some methods provide explanations about policy-level behaviors, while others explain specific, local decisions (e.g., ``Why does the agent select this but not that action in a state?'').
Although this work seeks to explain the agent's advice for the current state, we do not restrict to local explanations. The proposed {\name} approach provides a policy-level explanation that shows what the agent's advice would be in different states with varying features, which can help the human decision-maker better understand the agent's behavior, rather than providing a local explanation about the advised action only.  
Thus, {\name} intrinsically aims at increasing the humans' trust in the adviser agent (cf.~\cite{shin2021exptrust}).

Various types of policy-level explanations have been developed in prior works. 
For example, a video highlighting the agent's trajectories with important states is proposed in~\cite{amir2018highlights};
such trajectory summaries are augmented with saliency maps in~\cite{huber2021local}.
Abstracted policy graphs (i.e., Markov chains of abstract states) are introduced in~\cite{topin2019generation} for summarizing RL policies.
A chart illustrating the agent coordination and task ordering is used for policy summarization of multi-agent RL in~\cite{boggess2022toward}.  
Additionally, policy-level contrastive explanations (e.g., ``Why does the agent follow this but not that policy?'') have been considered in~\cite{sreedharan2022bridging,finkelstein2022deep,boggess2023explainable}.
 
To the best of our knowledge, however, none of the existing XRL methods uses a heatmap of domain-specific indices to summarize DRL input features as in {\name}. 
Furthermore, we overlay the heatmap with arrows visualizing (advised) optimal actions based on a trained RL policy and annotate these arrows with different shades of gray to indicate the importance degrees of states. 
We follow the notion of \emph{state importance} originally proposed in~\cite{torrey2013teaching}, which was adopted in~\cite{amir2018highlights} for summarizing the RL agent's behavior in a selected set of important states. 
By contrast, our explanation shows the agent's action in every state but highlights importance states with darker arrows.

\subsection{Explainable Recommendations}

There is a related line of work on \emph{explainable recommendations}~\cite{zhang2020explainable,vultureanu2022survey,naiseh2020explainable}, which refers to recommendation algorithms that not only provide recommendation results, but also explanations to clarify why such items are recommended.
For example, image and text-based explanations are generated in~\cite{yan2023personalized} by first selecting a personalized image set that is the most relevant to a user's interest toward a recommended item and then producing natural language explanations. 
User needs for explanations of recommendations are investigated in~\cite{tran2023user}, where studies find that users in high-involvement domains (e.g., selecting a car to buy) focus more on explanations compared to lower-involvement domains (e.g., selecting a movie to watch).

This work seeks to explain the agent's advice, which can be considered as a type of recommendation; and {\name} also provides explanations with both visual and textual information.
However, our problem setup is different from those recommendation algorithms, which usually do not consider repeated decision-making in complex environments.

\section{Problem Setup} \label{sec:problem} 
We consider a problem setup where an intelligent agent comprising a neural network-based prediction component and a deep reinforcement learning (DRL) component provides advice to a human decision-maker in complex repeated decision-making environments.

As illustrated in \autoref{fig:problem}, at each step, the agent makes some prediction $\hat{y}$ about the future based on the current state $s$ and historical data $h$, and generates an advice $a$ based on a trained DRL policy with the input $s$ and $\hat{y}$, and reward $r$;
the human decision-maker takes an action $a'$ where $a' = a$ if the human follows the agent's advice.  
But sometimes, an alternative action ($a' \neq a$) may be chosen if the human does not trust the agent or does not understand why the agent proposes a certain advice. 

This work aims to tackle this problem by generating explanations about the adviser agent to improve the human's trust and decision-making. 
We make two important assumptions as follows. 
\begin{itemize}
    \item \textbf{A1:} The agent is rational (i.e., seeking to maximize the expected discounted return) and not adversarial to the human decision-maker (i.e., no deception). 
    \item \textbf{A2:} The environment is based on a grid representation with discrete states and actions.  
\end{itemize}

\subsection{Motivating Examples} \label{sec:motiv}

The aforementioned problem setup is commonly shared by many complex repeated decision-making environments. 
Here we describe two motivating examples used in this work. 

\paragraph{Taxi environment}
Consider a taxi moving around in a grid world. In our example scenario, we assume the grid size to be $20\times20$. The taxi can stay put or 
move horizontally, vertically, or diagonally by up to two grid cells at each step; we assume that one step corresponds to ten minutes real time. The taxi receives a reward of $10$ for dropping off a passenger and a penalty of $-1$ per step for driving without any passenger. 
In each episode, the taxi starts at a random grid cell and time and terminates by the end of a nine-hour shift (i.e., $54$ steps). 

At each step of an episode, the adviser agent predicts the number of pick-up requests in each grid cell for the next step, based on a rich set of features, including the number of pick-up requests of the last 40 minutes, points of interest in each cell, as well as location-independent features such as date, time, holiday, and weather. 
Then, the agent advises an action for the taxi based on a DRL policy trained using the number of predicted pick-up requests and available taxis in each grid cell and the received reward.   

The taxi driver decides whether to follow the agent's advice or take an alternative action, which would impact the environment's feedback of state and reward. 
The above process repeats until the end of an episode. 

\paragraph{Wildfire environment}
Consider an aerial vehicle (AV) flying over a forest (modeled as a grid world) aiming to extinguish a wildfire. 
At each step (corresponding to 2.5 minutes), the AV can choose one of three types of actions: (1) extinguish the fire in the current grid cell, (2) stay put, or (3) decide to relocate by one cell in either of the four cardinal directions.
When the AV chooses the \textit{extinguish} action in a grid cell with a high neighborhood fire ratio, the AV receives a large positive reward that is calculated based on the neighborhood fire ratio. 
The AV receives a penalty of $-2.5$ for taking the \textit{extinguish} action in a grid cell with a low neighborhood fire ratio, and a cost of $-1$ per step for moving around. 
In each episode, the AV starts at a random grid cell and terminates after 100 steps. 

At each step, the adviser agent predicts the fire risk (i.e., the probability of fire occurrences) in each grid cell, based on features including each grid cell's forest fuel level and burning status. Then, the agent advises an action for the AV based on a DRL policy trained using the fire risk prediction, the current state and the received reward. 

The AV operator decides whether to follow the agent's advice, which would also affect the state of the environment.
The above process repeats until an episode terminates.

\subsection{Baseline Explanations} \label{sec:baseline}

The baseline explainer considers a black-box model consisting of the adviser agent's two components as a whole and explains input features' influence on the output advice. 
We apply LIME~\cite{ribeiro2016should} to check what happens to the agent's advice when the input features are perturbed and compute an influence value for each feature.  
We select LIME to generate baseline explanations (cf. \autoref{sec:baseline}), because SHAP is too slow for computational experiments. SHAP yields time-out (i.e., more than two minutes) for most models used in our experiments, while LIME and the proposed {\name} approach can generate explanations within a few seconds.
The generated baseline explanations are represented as saliency maps showing how much each feature contributes to the agent's advice. 

For an example saliency map illustrating the influence of each grid cell's current pick-up request counts on the agent's advice see the Appendix.
The baseline explanation generated at each step may include multiple saliency maps corresponding to different features.
For example, there are five saliency maps generated for the taxi environment per step.
We hypothesize that such a baseline explanation is overwhelming and cannot effectively assist humans with decision-making.

\section{Approach} \label{sec:approach} 

\begin{figure}[t]
\centering
\includegraphics[width=.5\textwidth]{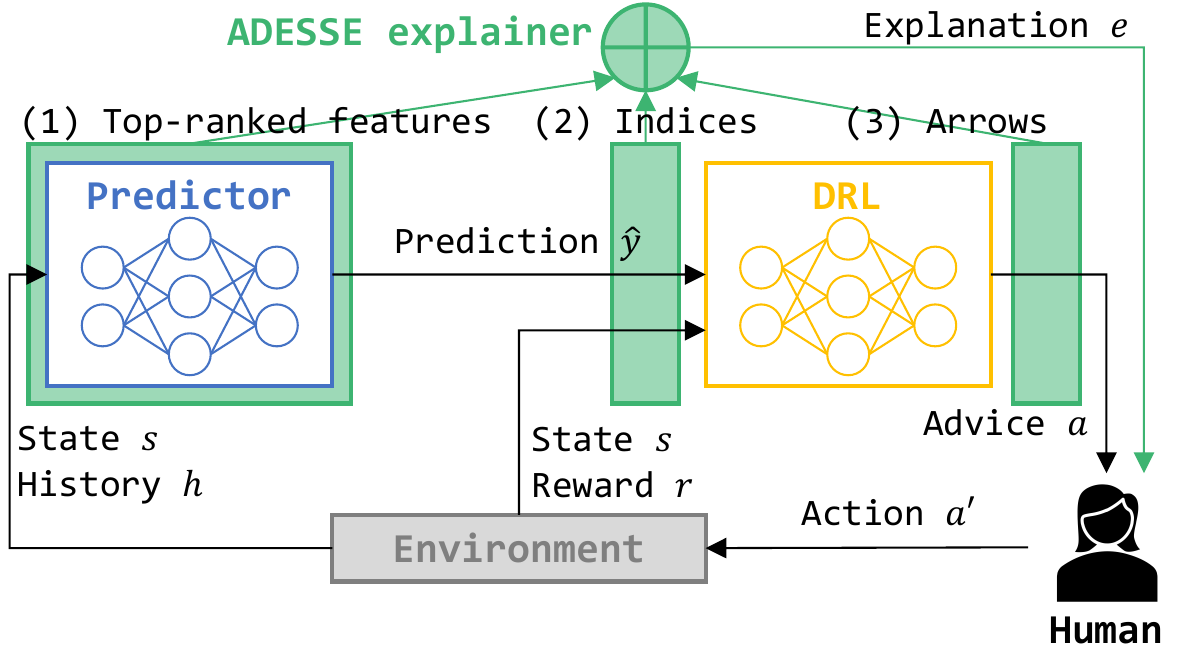}
\caption{An explanation generated by {\name} consists of (1) a list of top-ranked features for the prediction, (2) domain-specific indices summarizing the DRL input features, and (3) arrows visualizing the trained DRL policy.}
\label{fig:proposed}
\end{figure}

To address the limitations of baseline explanations, we propose an approach named {\name} that leverages the problem structure and generates explanations consisting of three key elements as shown in \autoref{fig:proposed}.
First, a list of top-ranked features is selected based on their contributions to the prediction (cf. \autoref{sec:shap}). 
Second, a domain-specific index function is used to summarize the DRL input features (cf. \autoref{sec:index}).
Third, the trained DRL policy is visualized as arrows in a grid world with importance degrees (cf. \autoref{sec:arrows}).
And finally, we describe how {\name} generates an explanation integrating these elements (cf. \autoref{sec:alg}). 

\subsection{Top-Ranked Features for the Prediction} \label{sec:shap}

We rank input features of the prediction component based on Shapley values computed via SHAP~\cite{lundberg2017unified}, which tells us the contribution of each feature to the prediction. 
We favor SHAP over LIME here, because identifying the top-ranked features for a few selected predictions with the smaller search space allows a fast computation time and we want the properties guaranteed by SHAP.

To reduce the explanation size, we focus on selecting a short list of top-ranked features for an individual prediction output at a time. 
For example, for the taxi environment, the human decision-maker may be interested to know what are the top six features contributing to the pick-up request prediction at the taxi's current location or the advised next location.
Such explanations could improve the human's trust in the agent's prediction component.

During a game-based user study (cf. \autoref{sec:study}), the human decision-maker can choose from a list of locations (e.g., grid cells labeled with A-F in \autoref{fig:proposed_taxi}) for displaying the top-ranked features that contribute the most to the prediction in each location. 

\subsection{Domain-Specific Indices} \label{sec:index}

To explain the input of the agent's DRL component, we summarize the DRL input features using a domain-specific index function, rather than showing multiple saliency maps (i.e., one for each DRL input feature) as in baseline explanations. 

\begin{figure}[t]
\centering
\includegraphics[width=.5\textwidth]{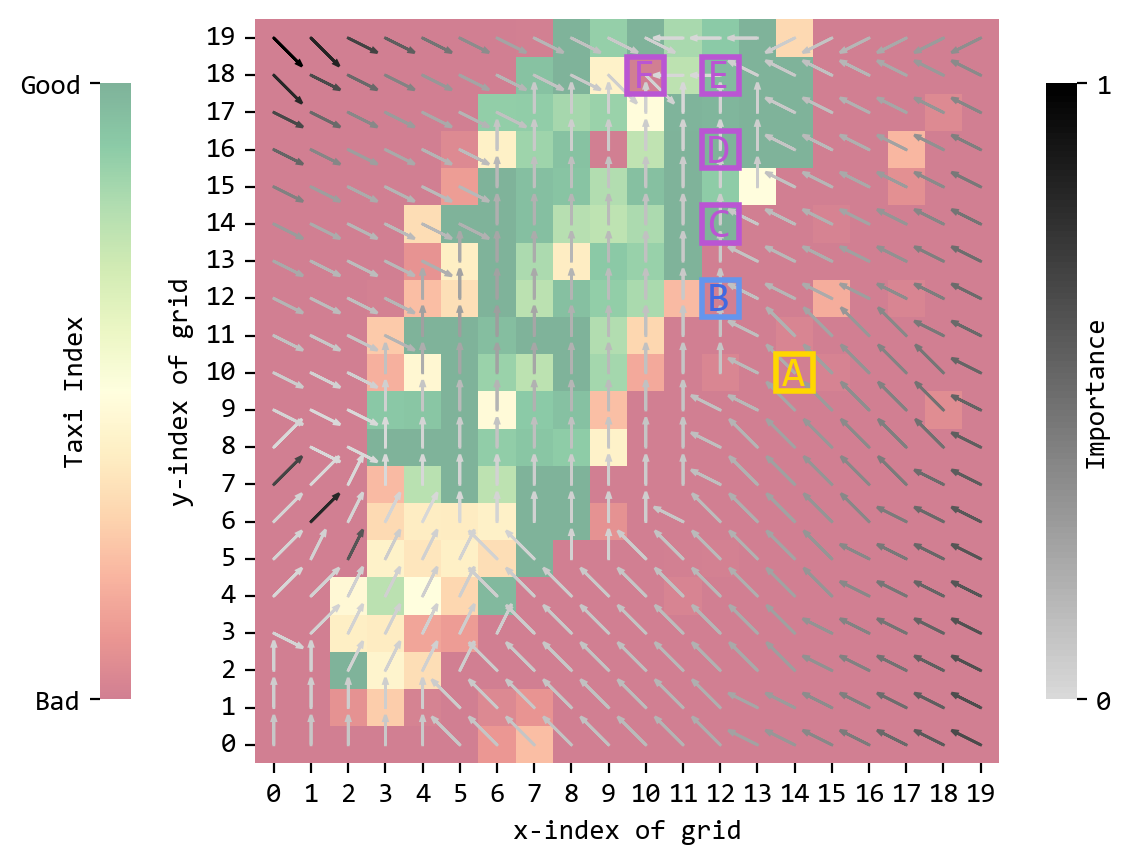}
\caption{An example of {\name} explanation for the taxi environment. A is the taxi's current location and B is the advised next location, which lead to C-F in the next few steps following the trained DRL policy visualized as arrows. A list of top-ranked features would be displayed separately when the human selects one of these labelled locations.}
\label{fig:proposed_taxi}
\end{figure}

\paragraph{Taxi environment}
Recall from \autoref{sec:motiv} that the DRL input features for the taxi environment include the number of predicted pick-up requests and available taxis in each grid cell. 
Inspired by the \emph{demand-supply ratio}, a metric commonly used in the market for taxi services~\cite{kamga2015analysis}, we define an index function for the taxi environment as follows. 

\begin{equation}
    \phi_{\mathsf{taxi}} (g) = 
    \begin{cases}
         \eta \cdot \frac{\rho(g)}{\tau(g)} + (1-\eta) \cdot \frac{\rho(g) \cdot |G|}{\sum_{g' \in G} \rho(g')} 
                & \textrm { if } \tau(g) > 0 \\
         0 & \textrm { otherwise }
    \end{cases}
\end{equation}
where $g \in G$ is a grid cell in the taxi grid world, $|G|$ is the total number of grid cells, and $\rho(g)$ and $\tau(g)$ are the number of predicted requests and available taxis in a grid cell $g$, respectively.  
We set $\eta =0.75$ to balance the trade-off between the demand-supply ratio of taxi services and the ratio of predicted requests in a grid cell $g$ compared with the average requests over the entire grid world $G$. 
\autoref{fig:proposed_taxi} shows an example heatmap of the obtained taxi indices where the darkest red indicates that $\phi_{\mathsf{taxi}}(g) = 0$.

\paragraph{Wildfire environment}
For each grid cell $g$ in the forest grid world $G$, the DRL input features for the wildfire environment include the predicted fire risk $\mu(g) \in [0,1]$, the normalized forest fuel level $\theta(g)\in [0,1]$ and the burning status $\beta(g) \in \{true, false\}$. 
We define an index function for the wildfire environment based on domain knowledge~\cite{haksar2018distributed,julian2019distributed} as follows.
\begin{equation}
    \phi_{\mathsf{fire}} (g) = 
    \begin{cases}
         - \theta(g)  \cdot \mu(g)  & \textrm { if } \beta(g) = true \\
         \left(1-\theta(g)\right) \cdot \left(1-\mu(g)\right) & \textrm { otherwise }
    \end{cases}
\end{equation}
The intuition is that, when a grid cell has caught fire, a higher forest fuel level and higher predicted fire risk would lead to more severe fire, and hence a more negative value of the wildfire index; conversely, when a cell is not on fire, it is safer (i.e., more positive value of the wildfire index) when there is a lower forest fuel level and lower predicted fire risk. 
An example heatmap of the obtained wildfire indices is shown in the Appendix.

\subsection{Arrows with Importance Degrees} \label{sec:arrows}

To improve the human decision-maker's trust in the adviser agent, we visualize the trained DRL policy for the entire grid world rather than only displaying the agent's advice for the current grid cell.  

Let $g_t \in G$ denote the current grid cell at time $t$. The DRL state $\sigma_t$ is given by the environment state $s_t$ (which includes $g_t$ as a feature) and the predicted state $\hat{y}_t$.
Let $\sigma_t[g]$ denote a DRL state that replaces $g_t$ with a grid cell $g \in G$ but preserves all the other features of $\sigma_t$. 
For example, in the taxi environment, $\sigma_t[g]$ represents a state where $g$ is an assumed location of the taxi, and the rest of DRL input features (i.e., number of predicted pick-up requests and available taxis at each grid cell) stay the same as in $\sigma_t$. 

Given a trained DRL policy $\pi_t$ at time $t$, the optimal action $a(g)$ in a grid cell $g$ seeks to maximize the Q-value that estimates the rewards ultimately achievable by taking an action in a state.
\begin{equation}
    a(g) = \pi_t(\sigma_t[g]) = \argmax_\alpha Q(\sigma_t[g], \alpha)
\end{equation}
where $\alpha$ denotes any possible action in state $\sigma_t[g]$. 

\autoref{fig:proposed_taxi} plots the optimal action in each grid cell as an arrow overlaying the index heatmap obtained from \autoref{sec:index}.
Moreover, we annotate these arrows with various shades of gray to represent the normalized importance degrees.
We define the importance degree of each grid cell $g \in G$ following the notion of \emph{state importance} proposed in~\cite{torrey2013teaching}: 
\begin{equation}
    I(g) = \max_\alpha Q(\sigma_t[g],\alpha) - \min_\alpha Q(\sigma_t[g],\alpha)
\end{equation}
Intuitively, if all actions in a state share the same Q-value, then the state is the least important for advising because it does not matter which action is chosen.  
We normalize importance degrees $I(g)$ over the entire grid world $G$ and obtain:
\begin{equation}
    \delta(g) = \frac{I(g) - \min_{g \in G} I(g)}{\max_{g \in G} I(g) - \min_{g \in G} I(g)}
\end{equation}
such that the normalized importance degree $\delta(g) \in [0,1]$. 

\subsection{Explanation Generation Algorithm} \label{sec:alg}

Algorithm~\ref{alg:proposed} illustrates the procedure of {\name} generating an explanation at a time step $t$ by integrating these aforementioned elements. 
First, a set of locations along a finite path starting from the current grid $g_t$ and following the trained DRL policy $\pi_t$ is identified (e.g., A-F in \autoref{fig:proposed_taxi}) and a list of top-ranked input features for the prediction in each location is selected as described in \autoref{sec:shap}. 
Next, for each grid $g \in G$ in the grid world, a domain-specific index (e.g., $\phi_{\mathsf{taxi}} (g)$ and $\phi_{\mathsf{fire}} (g)$ introduced in \autoref{sec:index}) is computed to summarize the DRL input features and plotted in a heatmap. 
Lastly, the optimal action $a(g)$ and the normalized importance degree $\delta(g)$ for each grid $g \in G$ is computed following \autoref{sec:arrows}, which are plotted as arrows with various shades of gray overlaying the heatmap of indices. The generated explanation is returned as a heatmap as shown in \autoref{fig:proposed_taxi}, together with separate lists of top-ranked features for the prediction. 

\begin{algorithm}[tb]
    \caption{Generating an explanation at a time step $t$}
    \label{alg:proposed}
    \textbf{Input}: Grid world $G$, current grid $g_t$, current state $s_t$, 
        predication input $x_t$ and output $\hat{y}_t$, 
        DRL input $\sigma_t \subseteq s_t \cup \hat{y}_t$ and policy $\pi_t$\\
    \textbf{Parameter}: Optional list of parameters\\
    \textbf{Output}: Explanation $e_t$
    \begin{algorithmic}[1] 
        \FORALL{$g$ in a finite path starting from $g_t$ following $\pi_t$}
            \STATE $F \gets $ append top-ranked features $f(g) \subset x_t$ \;
        \ENDFOR
        \FORALL{$g \in G$}
            \STATE Compute domain-specific indices $\phi(g)$ based on $\sigma_t$ \;
            \STATE Compute optimal action $a(g)$ \;
            \STATE Compute the normalized importance degree $\delta(g)$ \;
        \ENDFOR
        \STATE \textbf{return} $e_t = \langle F, \{\phi(g)\}_{g \in G}, \{a(g), \delta(g)\}_{g \in G}\rangle$
    \end{algorithmic}
\end{algorithm}

\section{Computational Experiments} \label{sec:exp} 
We build a prototype implementation\footnote{\href{https://github.com/sorensc/ADESSE}{https://github.com/sorensc/ADESSE}} of {\name} and compare its performance with the baseline explainer using LIME (cf. \autoref{sec:baseline}) via computational experiments on the taxi and wildfire environments with varying model sizes.

\subsection{Implementation}

\paragraph{Taxi environment}
We implemented the prediction component as a feed-forward neural network consisting of five fully connected layers with 20, 128, 64, 32, and 16 neurons; 
and utilized the dueling double deep Q-learning~\cite{wang2016dueling} for the DRL component (three convolutional and three fully connected layers). 
The New York City Yellow Taxi dataset\footnote{\href{https://www.nyc.gov/site/tlc/about/tlc-trip-record-data.page}{https://www.nyc.gov/site/tlc/about/tlc-trip-record-data.page}} was used for training and validation (186 million trips taken between January 2015 and June 2016), where the GPS start and end locations of trips were mapped to grid cells in the environment.

\paragraph{Wildfire environment}
For this environment, we implemented the prediction component as a feed-forward neural network with three layers of 6, 512, and 512 neurons; 
as in the taxi environment, dueling double deep Q-learning~\cite{wang2016dueling} was used for the DRL component (three convolutional and three fully connected layers). 
The environment dynamics (forest fire model) was adapted from~\cite{haksar2018distributed,julian2019distributed}.

\paragraph{Setup}
All experiments were run on a MacBook laptop with an Apple M1 Pro chip, 32 GB of memory, and Ventura 13.5.2 operating system.

\subsection{Results}

\begin{table}[t]
\begin{center}
\begin{tabular}{crrrrr} \toprule
\multicolumn{2}{c}{Environment} & \multicolumn{2}{c}{Explanation Size} & \multicolumn{2}{c}{Time (seconds)} \\
  \cmidrule(r){1-2}   \cmidrule(r){3-4} \cmidrule(r){5-6} 
Domain & $|G|$ & Baseline & {\name} & Baseline & {\name} \\
\midrule
Taxi   & 10$\times$10 & 0.71K & \textbf{0.24K} &  7.2 & \textbf{0.9} \\
       & 20$\times$20 & 2.81K & \textbf{0.84K} &  10.0 & \textbf{1.3} \\    
       & 40$\times$40 & 11.21K & \textbf{3.24K} &  18.3 & \textbf{5.9} \\  
       & 80$\times$80 & 44.81K & \textbf{12.84K} &  41.2 & \textbf{25.5} \\ 
\midrule
Wildfire   & 10$\times$10 & 0.30K & \textbf{0.24K} &  0.9 & \textbf{0.2} \\
           & 20$\times$20 & 1.20K & \textbf{0.84K} &  1.5 & \textbf{0.4} \\    
           & 40$\times$40 & 4.80K & \textbf{3.24K} &  2.9 & \textbf{0.9} \\  
           & 80$\times$80 & 19.20K & \textbf{12.84K} &  8.5 & \textbf{3.0} \\ 
\bottomrule
\end{tabular}
\caption{Results of computational experiments.}
\label{tab:exp}
\end{center}
\end{table}

\autoref{tab:exp} shows the experimental results. 
For each model, we report the grid world size $|G|$, and compare the baseline and {\name} in terms of
the explanation size and the average time of generating an explanation per step over 10 independent runs. 
We draw the following key insights from the results: 
\begin{itemize}
    \item Both the baseline explainer and {\name} can successfully generate explanations for different environments with varying model sizes.  
    \item The size of {\name} explanation is significantly smaller than that of the baseline explanation across all models, and the size difference increases as the grid world grows larger. 
    \item {\name} is generally faster than the baseline and can generate an explanation within a few seconds for all models used in the experiments.   
\end{itemize}

\section{Game-Based User Study} \label{sec:study} 

We evaluate the effectiveness of explanations generated by {\name} via an interactive game-based user study\footnote{The study was approved by institutional review board.}.
We describe the study design in \autoref{sec:design}, report the results and discuss the insights in \autoref{sec:results}. 

\subsection{Study Design} \label{sec:design}

\paragraph{Game}
We designed an interactive game based on the taxi environment described in \autoref{sec:motiv}.
Each study participant was asked to act as a taxi driver who was incentivized to choose the optimal action in the environment at each step, in order to receive a high reward. 
The participants were presented with baseline explanations and with explanations generated by {\name}; our goal was to find to what extent and how this influences their decisions in terms of whether or not to follow the advised actions.
An example screenshot of the game user interface is shown in the Appendix.

\paragraph{Participants}
We recruited 28 participants; all of them were over the age of 18, fluent in English (since the game instructions were written in English), and did not have color blindness (which would have affected their ability to recognize the presented explanations). 
The average age of the participants was 28.96 years with a standard deviation of 8.27 years\footnote{Note that drivers of private transportation services such as Uber represent the demographic group from which we recruited the subjects for the study.}. 
39\% of the participants were female and 61\% male.
To ensure data quality, each participant responded to three attention-check questions during the study. 

\paragraph{Independent variables}
We adopted a within-subject study design where participants were asked to engage in two study trials, each of which involved playing the game for twelve steps with explanations generated by 
either {\name} or by the baseline explainer using LIME.
To counterbalance the ordering confound effect, one half of the participants were randomly selected to start the study trial with baseline explanations, followed by a trial with explanations generated by {\name}; the other half of the participants took the two study trials in reversed order.   

\paragraph{Dependent variables}
We recorded the average time spent to choose an action, the total reward achieved in a study trial, and the percentage of steps where the agent's advice was followed in a trial. 
At the end of each study trial, we also collected the participant ratings on a 5-point Likert scale (1 - strongly disagree, 5 - strongly agree) about the following statements adapted from~\cite{hoffman2018metrics} regarding \textbf{\emph{the explanation satisfaction scale}}:
\begin{itemize}
    \item The explanations help me \emph{understand} how the agent's advice is computed. 
    \item The explanations are \emph{satisfying}.
    \item The explanations are sufficiently \emph{detailed}.
    \item The explanations are sufficiently \emph{complete}, that is, they provide me with all the needed information to make decisions.
    \item The explanations are \emph{actionable}, that is, they help me know how to make decisions.
    \item The explanations let me know how \emph{reliable} the agent is for decision support.
    \item The explanations let me know how \emph{trustworthy} the agent is for decision support.
\end{itemize}

\paragraph{Procedure}
During the study, each participant was first briefed about the study purpose and the game instructions.  
Then, the participant was asked to play a study trial with one type of explanation (i.e., baseline or {\name}) and give ratings on the explanation satisfaction scale. 
Next, the participant was asked to play a second trial with the other explanation type, followed by explanation satisfaction ratings. 
The study was wrapped up with demographic questions (e.g., age, gender).
Additionally, to gain better insights into the behavior of participants, we asked a randomly selected set of participants to describe what their decision-making strategy was, and give an appraisal of how confident they were to choose a better action than the agent's advice.

\paragraph{Hypotheses}
We investigated three hypotheses stated below. 
\begin{itemize}
    \item \textbf{H1:} Explanations generated by {\name} \emph{lead to higher ratings on the explanation satisfaction scale} than the baseline.  
    \item \textbf{H2:} Explanations generated by {\name} enable the participants to \emph{take less time to choose actions} than the baseline. 
    \item \textbf{H3:} Explanations generated by {\name} enable the participants to \emph{achieve a higher total reward} than the baseline.    
\end{itemize}

\subsection{Study Results and Discussion} \label{sec:results}

We utilized a Wilcoxon signed-rank test to evaluate H1 and used a paired t-test to evaluate H2 and H3. 
For all tests, we set the significance level as $0.05$.

\begin{figure}[t]
    \begin{centering}
    \includegraphics[width=.5\textwidth]{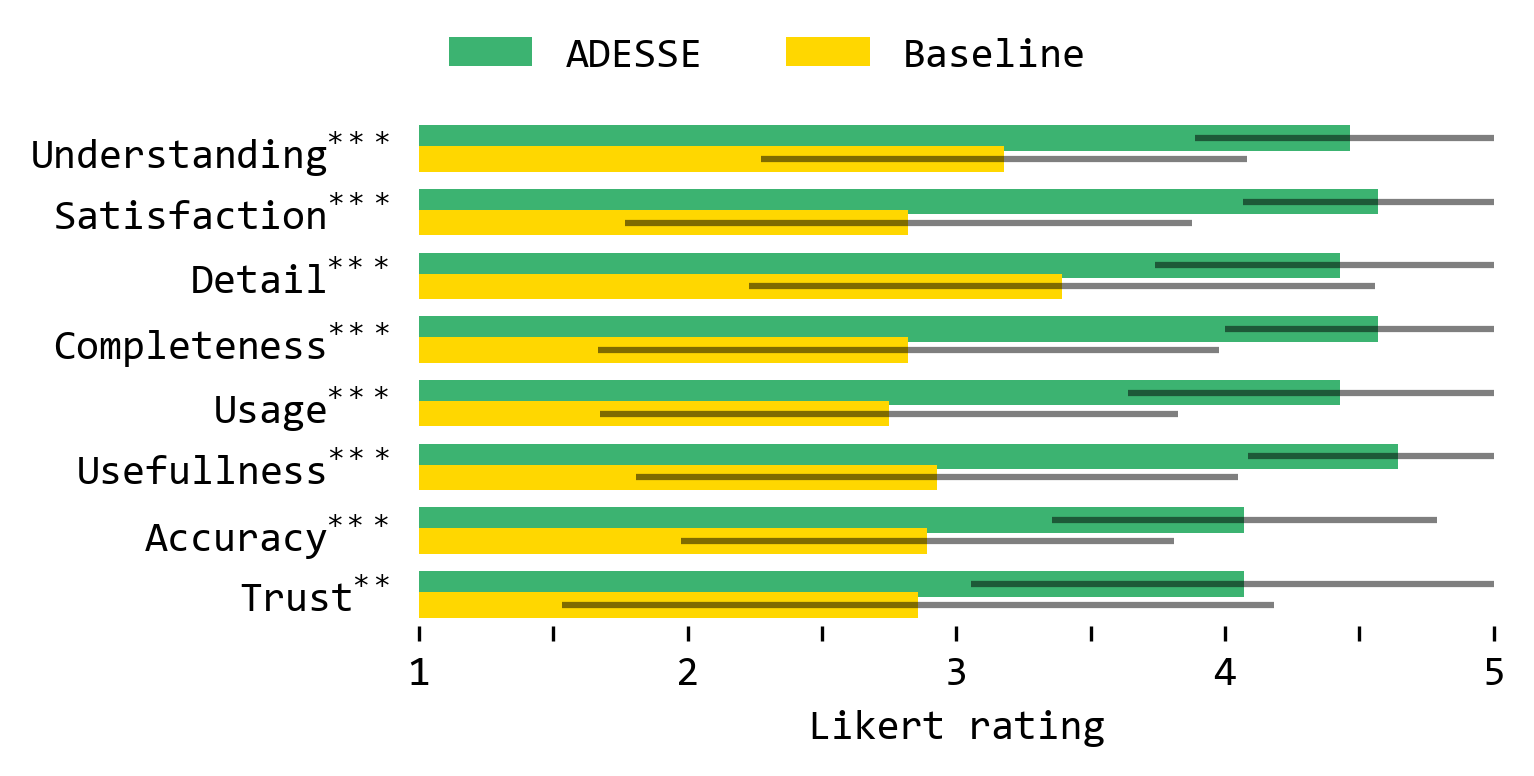}
    \caption{Mean and standard deviation of participant ratings on the explanation satisfaction scale comparing explanations generated by {\name} (top/green) and the baseline (bottom/gold). 
    ($^{**}$ for $0.001 < p \leq 0.01$ and $^{***}$ for $p \leq 0.001$.)}
    \label{fig:SatisfactionScale}
    \end{centering}
\end{figure}

\paragraph{Explanation satisfaction scale ratings}
As shown in \autoref{fig:SatisfactionScale}, participants ratings of explanations generated by {\name} are higher than ratings of the baseline in all explanation satisfaction scale metrics with statistically significant differences.
\emph{Thus, the data supports H1.}

\paragraph{Time for choosing actions}
On average, participants took less time to choose actions when being presented with explanations generated by {\name} ($M=38.78$, $SD=15.90$) compared to baseline explanations ($M=52.82$, $SD=27.72$). The difference is statistically significant ($t=2.9182$, $p=0.0070$). \emph{Thus, the data supports H2.}

\paragraph{Total reward}
The participants achieved a higher average reward when being presented with explanations generated by {\name} ($M=98.18$, $SD=13.18$) than baseline explanations ($M=90.18$, $SD=18.13$).
However, the paired t-test yields ($t=-1.8216$, $p=0.0796$) with the $p$ value slightly higher than $0.05$. \emph{Thus, the data partially supports H3.} 

\paragraph{Discussion}
One of the reasons that participants were more satisfied with explanations generated by {\name}, as indicated by the higher ratings on the explanation satisfaction scale, could due to the fact that explanations generated by {\name} are more succinct and informative than baseline explanations (note that there are five saliency maps in each baseline explanation generated by LIME). 
This may also justify the reason of participants took less time to choose actions with explanations generated by {\name}, since it requires more time to read and understand baseline explanations.

\section{Conclusion} \label{sec:conclu} 
We presented {\name}, a novel approach for generating visual and text-based explanations about an intelligent agent that provides advice to a human decision-maker in complex repeated decision-making environments. 
The agent consists of two deep learning-based components: one for making predictions about the future, and the other for computing advised actions with deep reinforcement learning based on the predicted future and the current state. 
{\name} leverages the agent's two-component structure and generates explanations with visual and textual information, to improve the human's trust in the agent and thus better assist human decision-making. 
Results of computational experiments demonstrate the applicability and scalability of {\name}, while an interactive game-based user study shows the effectiveness of explanations generated by {\name}.

There are several directions to explore for possible future work. 
First, we will extend {\name} to be able to deal with environments with continuous state/action space, beyond grid world environments considered in this work. 
For example, there has been increasing interest in using deep learning to predict future blood glucose levels of diabetes patients and then compute an advised insulin dosage based on the prediction via deep reinforcement learning~\cite{emerson2023offline}. 
Moreover, we will explore an extension to the multi-agent setting where advice is computed via multi-agent DRL.

\section*{Acknowledgements}
Sören Schleibaum was supported by the Deutsche Forschungsgemeinschaft under grant 227198829/GRK1931.
Lu Feng was supported by U.\,S. National Science Foundation under grant CCF-1942836. Sarit Kraus was supported in part by the EU Project TAILOR under grant 952215. 

\bibliographystyle{unsrt}  
\bibliography{references}  

\newpage
\begin{appendices}
\label{sec:Appendix}
Here, we describe details about the repositioning agent in \autoref{app:DetailsOfRepositioningAgent}, the wildfire agent in \autoref{app:DetailsOfWildfireAgent}, and provide more information about the user study in \autoref{app:UserStudyDetails}. Besides this document, the technical appendix contains
\begin{enumerate*}[label=(\arabic*)]
    \item a file \path{game_log.csv} that contains the logs of each study participant playing the game with the baseline and our composed setting, and
    \item the corresponding responses to the questionnaire in \path{questionnaire_response.csv}.
\end{enumerate*}
Upon acceptance of the paper, the source code will be published.

\section{Details of the Repositioning Agent}
\label{app:DetailsOfRepositioningAgent}

Here, we describe the dataset, the request estimation, and the repositioning in more detail.

\subsection{Dataset} 
\begin{figure}[ht]
    \centerline{\includegraphics[width=.5\textwidth]{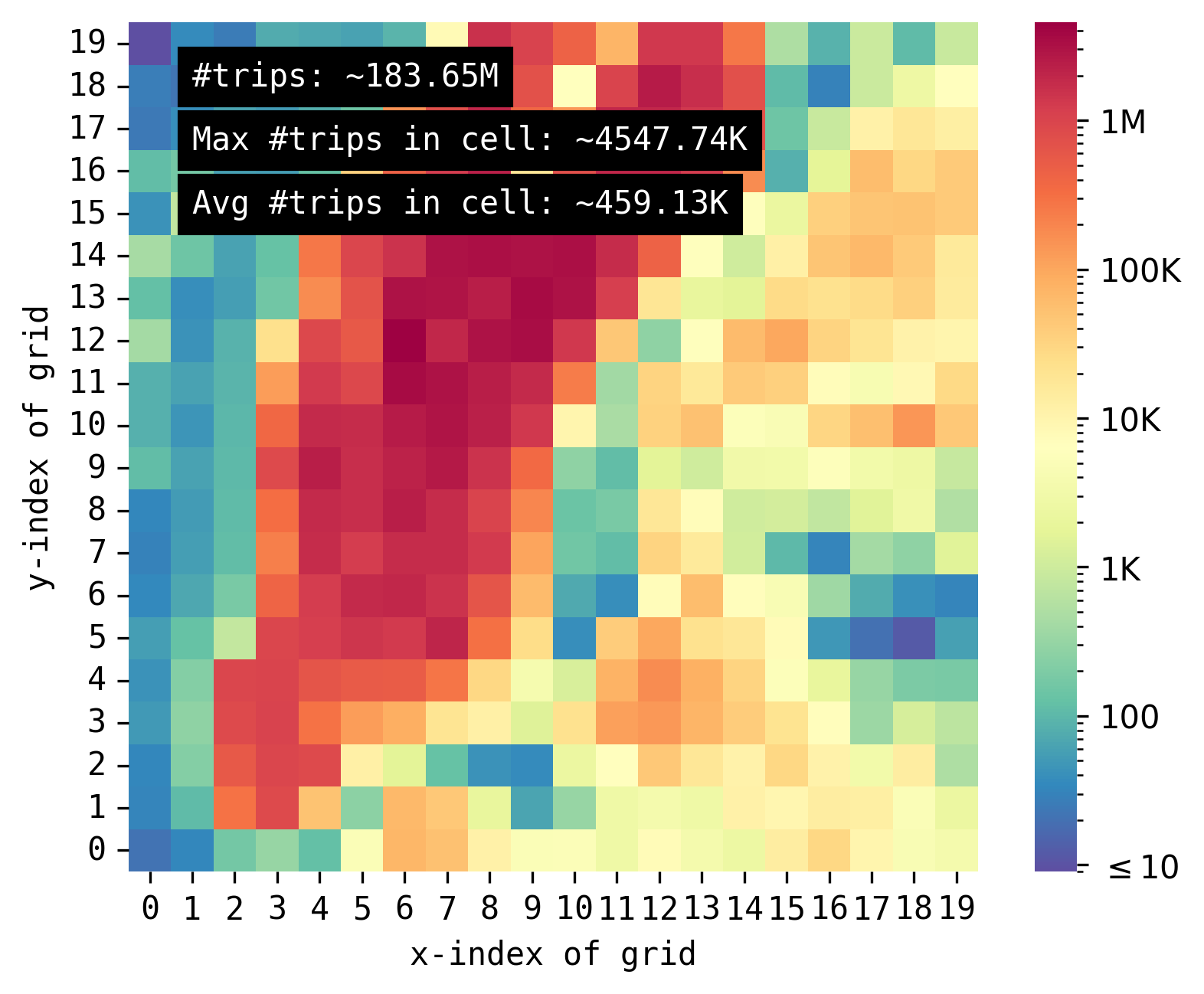}}
    \caption{Distribution of the number of taxi trips in the New York City Yellow Taxi dataset in 2015 and 2016 visualized on a logarithmic scale through a 500m square grid.}
    \label{fig:TripDistribution}
\end{figure}

In \autoref{fig:TripDistribution}, we visualize the distribution of the pick-up locations of the New York City Yellow Taxi dataset on a logarithmic scale. We know that this demand for requests is not the real one as it only includes those trips that took place. Nevertheless, we stick to the this dataset as there is no alternative with such a large number of trips publicly available.

\subsection{Request Prediction} 
The request predictor consists of five fully connected layers with 20, 128, 64, 32, and 16 neurons. With a learning rate of $0.001$ and $15$ epochs of training, we achieve a mean absolute error of $1.26$ trips per cell on the test data. As input features, we used the following:
\begin{enumerate*}[label=(\arabic*)]
    \item x-index at location $v$,
    \item y-index at $v$,
    \item \#requests 30 minutes ago at $v$,
    \item \#requests 20 minutes ago at $v$,
    \item \#requests 10 minutes ago at $v$,
    \item \#requests now at $v$,
    \item \#points of interests at $v$,
    \item hour,
    \item minute,
    \item weekday,
    \item month,
    \item temperature,
    \item wind,
    \item humidity,
    \item air pressure,
    \item view,
    \item whether there is snow,
    \item precipitation,
    \item whether there are clouds, and
    \item whether the day is a public holiday.
\end{enumerate*}

\subsection{DRL-based Repositioning}
We train the repositioning agent via deep reinforcement learning. Similar to \cite{haliem.2021} and related work in taxi repositioning, we use model-free off-policy Q-learning. In particular, we use dueling double deep Q-learning as proposed by \cite{wang.2016} with experience replay as it is closer to the state of the art in reinforcement learning than the double deep Q-learning approach used by \cite{haliem.2021}. Both networks -- the policy and the target one -- consist of three convolutional layers with corresponding kernel sizes of 5, 5, and 3; the number of filters is set to 16, 32, and 64. The next layer is fully connected with 64 * 12 * 12 + 2 = 9218 input and 1024 output neurons. Next, the dueling approach differentiates between a value and an advantage layer, which both receive the same input from the previous layers. As we do not aim to outperform other repositioning approaches but to enable explaining them, we tune the hyperparameter manually, resulting in 
\begin{enumerate*}[label=(\arabic*)]
    \item a learning rate of 0.001,
    \item a gamma of 0.99,
    \item an episode decay of 675 to adjust the exploration-exploitation trade-off,
    \item a target network update rate of 11,
    \item and a replay memory size of 15K transitions.
\end{enumerate*} 
On average, the repositioner achieves an reward of 6.85 per step in the test data.

\section{Details of the Wildfire Extinguishing Agent}
\label{app:DetailsOfWildfireAgent}

To provide further details of the wildfire extinguishing agent, we provide details of the fire model, the fire prediction, the DRL-based wildfire extinguishing agent, and an illustration of the corresponding explanation generated by {\name}. 

\subsection{Fire Model} 
At each time step the forest is updated, which consists of three steps:
\begin{enumerate*}[label=(\arabic*)]
    \item When a cell is burning, its fuel is reduced by a value $\beta$.
    \item For each grid cell $g \in G$ the amount of cells burning in the neighborhood ($b_{g}$) is calculated. If $g$ is not burning and $b_g > v * \alpha$, the cell is set on fire.
    \item In case the aerial vehicle dropped water to extinguish the fire on burning cells, those stop burning.  
\end{enumerate*}
We set $\alpha = 20$ and $\beta = 0.7$.

\subsection{Fire Prediction} 
To predict the risk of fire at a location $g \in G$ the fire estimation model receives the
\begin{enumerate*}[label=(\arabic*)]
    \item forest fuel,
    \item whether or not the forest is burning, and the
    \item location of the aerial vehicle.
\end{enumerate*}
We use a feed-forward fully-connected neural network with three layers and 6, 512, and 512 nodes. As activation functions, we select Leaky ReLU and Sigmoid. The network is optimized via Adam with an initial learning rate of 0.01 through mean squared error loss. We train for up to 1000 steps, but apply early stopping with a patience of 100 steps. Further, we save the best model during training.

\subsection{DRL-based Wildfire Extinguishing}
The reward function used for training depends on the percentage of the fire extinguished, the number of forest cells extinguished, and the time step. We train the extinguishing agent via dueling double deep Q-learning with experience replay. Both networks -- the policy and the target one -- consist of three convolutional layers with corresponding kernel sizes of 3, 3, and 3; the number of filters is set to 4, 8, and 16. The fully connected layers have 68, 1024, 512, 256. Next, the dueling approach differentiates between a value and an advantage layer, which both receive the same input from the previous layers. As we do not aim to outperform other repositioning approaches but to enable explaining them, we tune the hyperparameter manually, resulting in 
\begin{enumerate*}[label=(\arabic*)]
    \item a learning rate of 0.001,
    \item a gamma of 0.95,
    \item an episode decay of 675,
    \item a target network update rate of 256,
    \item and a replay memory size of 15K transitions.
\end{enumerate*} 
After training, the agent is able to preserve 72.86\% of the original forest fuel from burning.

\subsection{Explanation Illustration}
\begin{figure}[t]
\centering
\includegraphics[width=.5\textwidth]{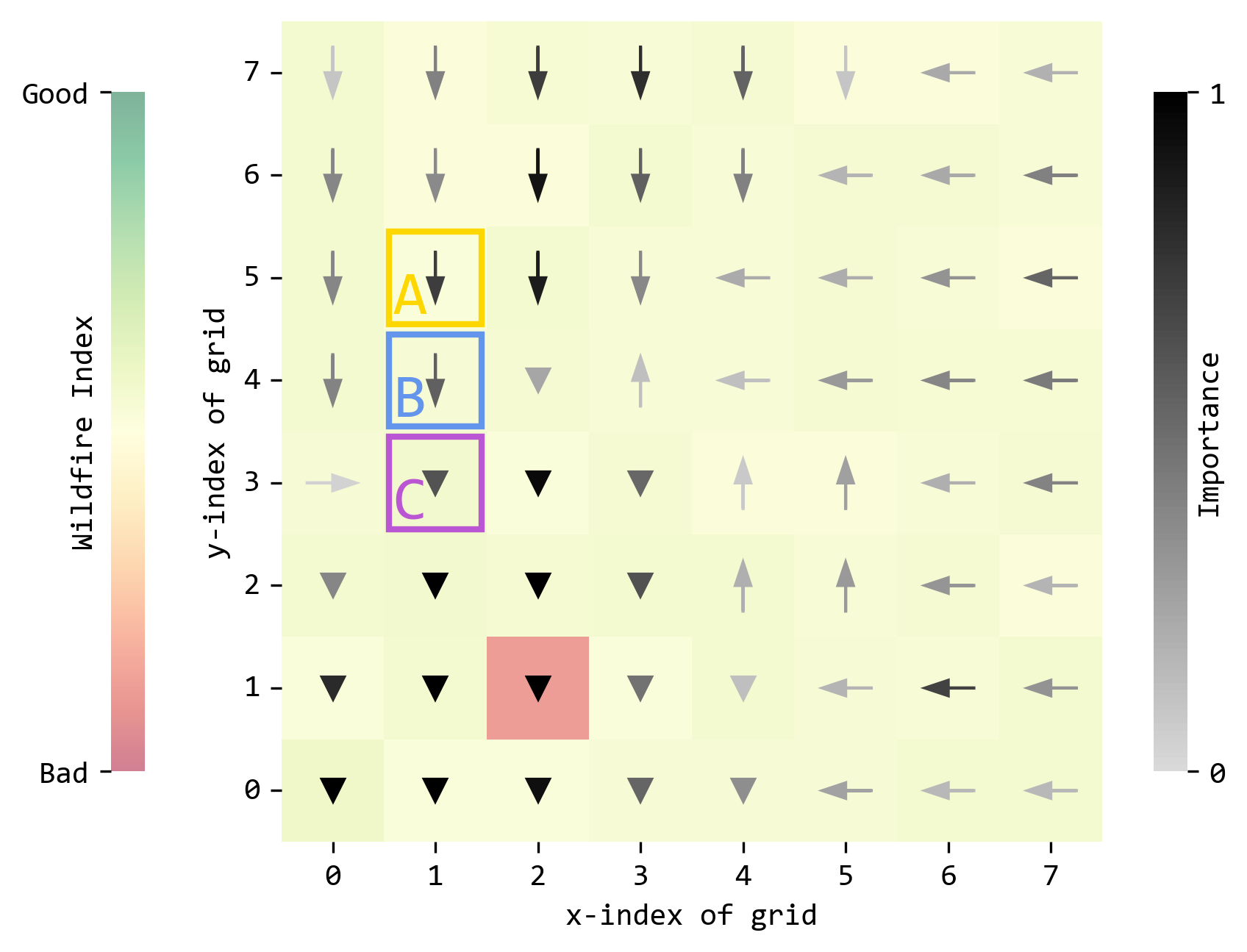}
\caption{An example of {\name} explanation for the wildfire environment. A is the AV's current location and B is the advised next location, which lead to C in the next step following the trained DRL policy visualized as arrows. A list of top-ranked features would be displayed separately when the human selects one of these labelled locations.}
\label{fig:proposed_wildfire}
\end{figure}

\autoref{fig:proposed_wildfire} shows an example heatmap of the obtained wildfire indices, where the darkest green and red indicate that $\phi_{\mathsf{fire}}(g) = 1$ and $\phi_{\mathsf{fire}}(g) = -1$, respectively.

\section{Details of User Study}
\label{app:UserStudyDetails}

Here, we describe the profile of the study respondents, report the description of the game given to the participants, show the graphical user interface of the game, and list the questions of the questionnaire conducted in the study.

\subsection{Profile of Respondents}
We list the profile of the study participants in \autoref{tab:ProfileOfRespondents}. We consider the reported sex, age, education, and country of living.

\begin{table}[t]
  \begin{centering}
	\begin{tabular}{llrr}\toprule
		\textbf{Question}                                 & \textbf{Answer}     & \multicolumn{2}{c}{\textbf{Overall ($n = 28$)}} \\
		\cmidrule(r){3-4} &                                                     & \textbf{n} &                         \textbf{\%} \\ \midrule
		Sex                                               & Female              &       $11$ &                                $39$ \\
		                                                     & Male                &       $17$ &                                $61$ \\
		                                                     & No gender           &          - &                                   - \\
                                                              & No answer           &          - &                                   - \\
		Age                                               & $< 21$              &        $3$ &                                $11$ \\
		                                                     & $21$ to $30$        &       $17$ &                                $61$ \\
		                                                     & $31$ to $40$        &        $5$ &                                $18$ \\
		                                                     & $41$ to $50$        &        $1$ &                                 $4$ \\
		                                                     & $51$ to $60$        &          - &                                   - \\
		                                                     & $> 60$              &        $1$ &                                 $4$ \\
		                                                     & No answer           &        $1$ &                                 $4$ \\
		Education                                         & No training yet     &          - &                                   - \\
		                                                     & Secondary school    &        $1$ &                                 $4$ \\
		                                                     & High school diploma &        $3$ &                                $11$ \\
		                                                     & Vocational training &        $2$ &                                 $7$ \\
		                                                     & Bachelor degree     &        $8$ &                                $29$ \\
		                                                     & Master degree       &       $11$ &                                $39$ \\
		                                                     & Doctorates          &        $3$ &                                $11$ \\
		                                                     & Other               &          - &                                   - \\
		                                                     & No answer           &          - &                                   - \\
		Country                                           & Germany             &       $18$ &                                $64$ \\
		                                                     & Israel              &        $6$ &                                $21$ \\
                                                              & United States       &        $3$ &                                $11$ \\
                                                              & Finland             &        $1$ &                                 $4$ \\
		                                                     & No answer           &          - &                                   - \\ \bottomrule
	\end{tabular}
    \caption{Profile of respondents}
    \label{tab:ProfileOfRespondents}
\end{centering}
\end{table}

\subsection{Description of Game Given to Participants}
Before each participant starts playing the game, we describe that she is a taxi driver aiming to maximize her reward. Further, we describe the following aspects: 
\begin{enumerate*}[label=(\arabic*)]
    \item the current location -- yellow square -- the advice -- blue square -- and the last location -- black square --
    \item that at each step a movement of up to two cells or staying at the current location is possible via the action buttons,
    \item the reward function,
    \item the available information fields like the accumulated reward, 
    \item the usage of the webpage (e.g., minimizing/maximizing of graphics and description pane) and
    \item the description of the explanation configuration.
\end{enumerate*}

\subsection{GUI of Game} 
\begin{figure*}[t]
\centering
\includegraphics[width=\textwidth]{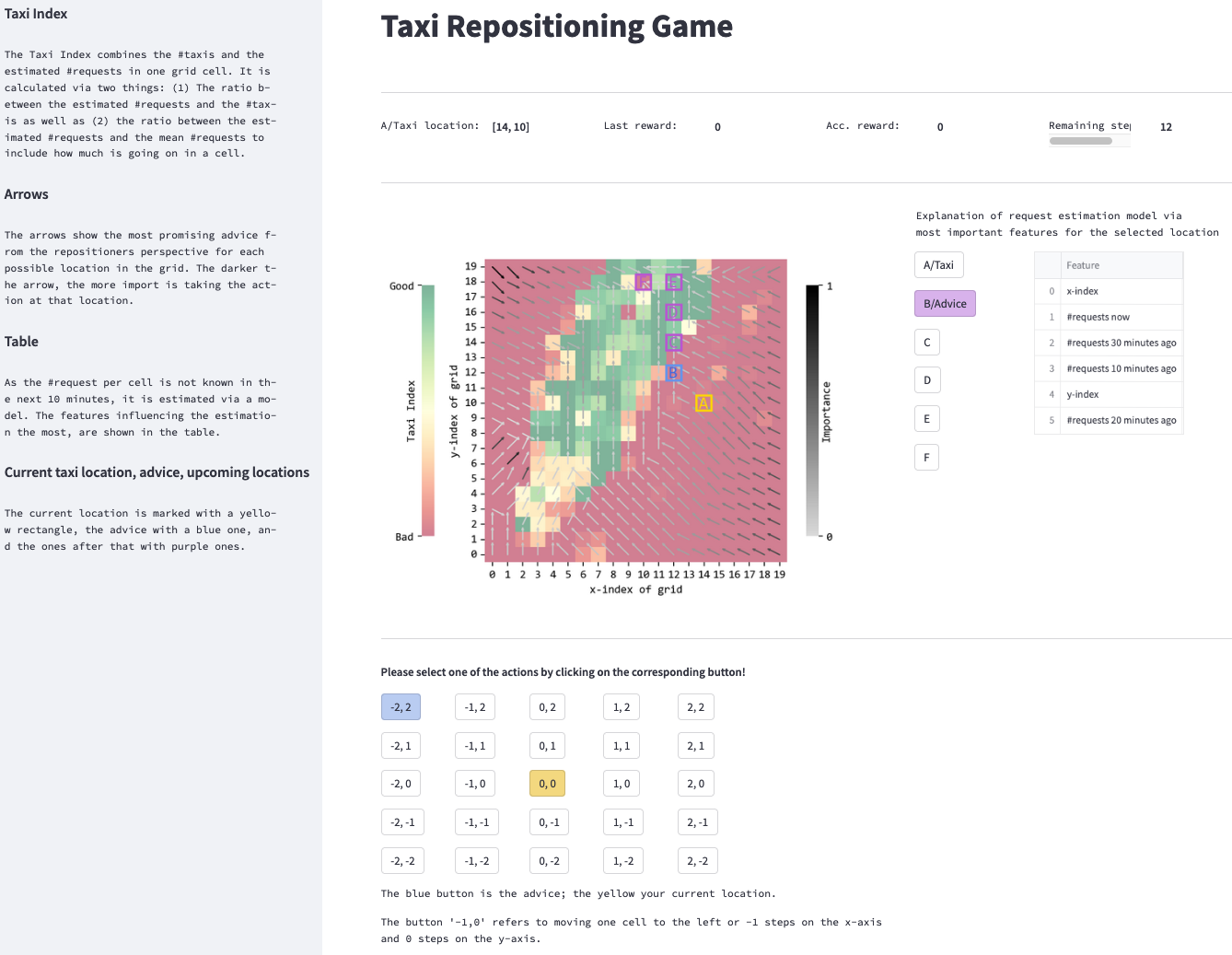}
\caption{An example screenshot of the game user interface where a participant is presented with an explanation generated by {\name} (top of the figure) and asked to select an action (bottom of the figure). The taxi's current location is shown in yellow and the agent's advice is shown in blue. The left panel reminds the participant about the game instructions.}
\label{fig:GameGUI}
\end{figure*}

\begin{figure*}
    \centerline{\includegraphics[width=\textwidth]{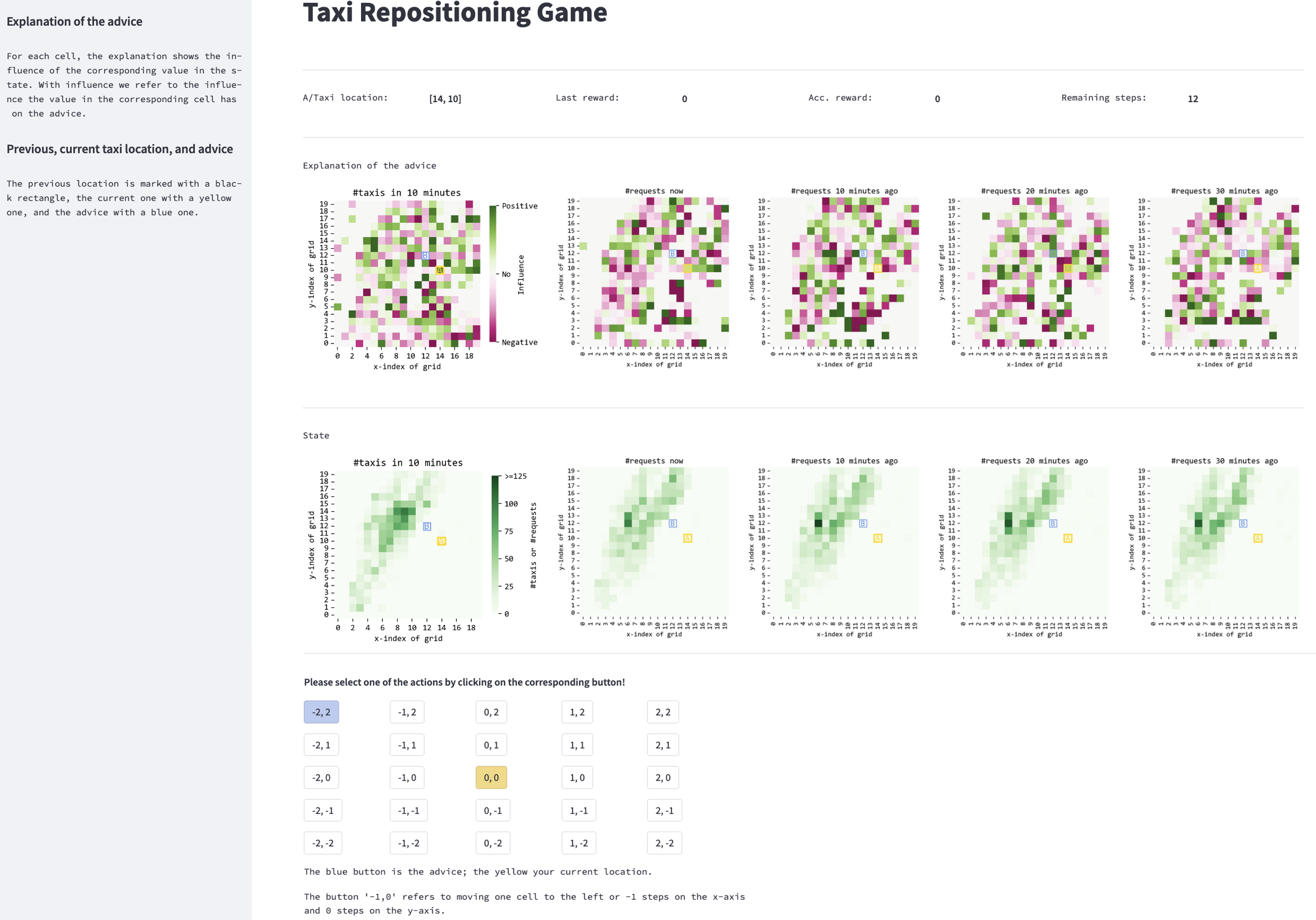}}
    \caption{An example screenshot of the game user interface where a participant is presented with an explanation generated by the baseline (top of the figure) and asked to select an action (bottom of the figure). The taxi's current location is shown in yellow and the agent's advice is shown in blue. The left panel reminds the participant about the game instructions.}
    \label{fig:GameGUIBL}
\end{figure*}

In \autoref{fig:GameGUI}, we show the graphical user interface of the game used in our study with our composed explanation; in \autoref{fig:GameGUIBL}, we show the corresponding counterpart for the baseline explanation.

\subsection{Questions}
Here, we list the questions, the participants answered during the game-based user study.

\paragraph{Step 1 - Introduction to the study and the game.}
\begin{itemize}
    \item Does the blue or the yellow square represent the location of the taxi? Queried via a multiple choice question.
    \item Does the blue or the yellow square represent the location of the advice? Queried via a multiple choice question.
    \item When not limited by the city borders, can you move up to two or four cells in each direction? Queried via a multiple choice question.
    \item Is the goal of the game to minimize or maximize the reward? Queried via a multiple choice question.
\end{itemize}

\paragraph{Step 3/5 - Questions related to the subjective usage of the advice.}
\begin{itemize}
    \item How much did you follow the given advices (marked with blue)? Queried on a 5-point Likert scale.
    \item In case you didn't follow the advices all the time: How confident are you, that you chose better than the given advices? Queried on a 5-point Likert scale.
    \item Please explain your strategy. Queried via an open text field.
\end{itemize}

\paragraph{Step 6 - Questions related to the explanation satisfaction scale.} The corresponding questions were already described in Section~6.

\paragraph{Step 7 - Demographic questions} 
\begin{itemize}
    \item How old are you?
    \item Which gender do you feel you belong to?
    \item What is currently your highest level of education?
    \item Which country do you live in?
\end{itemize}

\end{appendices}

\end{document}